\documentclass[letterpaper]{article} % DO NOT CHANGE THIS
\usepackage{aaai24}  % DO NOT CHANGE THIS
\usepackage{times,comment}  % DO NOT CHANGE THIS
\usepackage{helvet}  % DO NOT CHANGE THIS
\usepackage{courier}  % DO NOT CHANGE THIS
\usepackage[hyphens]{url}  % DO NOT CHANGE THIS
\usepackage{graphicx} % DO NOT CHANGE THIS
\usepackage{natbib}  % DO NOT CHANGE THIS AND DO NOT ADD ANY OPTIONS TO IT
\usepackage{caption} % DO NOT CHANGE THIS AND DO NOT ADD ANY OPTIONS TO IT
\usepackage{algorithm}
\usepackage{algorithmic}
\usepackage{multirow}

\usepackage{newfloat}
\usepackage{listings}
\DeclareCaptionStyle{ruled}{labelfont=normalfont,labelsep=colon,strut=off} % DO NOT CHANGE THIS
\floatstyle{ruled}
\newfloat{listing}{tb}{lst}{}
\floatname{listing}{Listing}
\nocopyright
\title{CFGPT: Chinese Financial Assistant with Large Language Model}
\author {
    Jiangtong Li\textsuperscript{\rm 1},
    Yuxuan Bian\textsuperscript{\rm 1},
    Guoxuan Wang\textsuperscript{\rm 1},
    Yang Lei\textsuperscript{\rm 1},
    Dawei Cheng\textsuperscript{\rm 1,2},\\
    Zhijun Ding\textsuperscript{\rm 1,2},
    Changjun Jiang\textsuperscript{\rm 1,2}
}
\affiliations {
    \textsuperscript{\rm 1} Department of Computer Science and Technology, Tongji University, Shanghai, China.  \\
    \textsuperscript{\rm 2} Shanghai Artificial Intelligence Laboratory, Shanghai, China.
}

\usepackage{bibentry}
\begin{document}

\maketitle

\begin{abstract}

Large language models (LLMs) have demonstrated great potential in natural language processing tasks within the financial domain.
In this work, we present a Chinese Financial Generative Pre-trained Transformer framework, named CFGPT, which includes a dataset~(CFData) for pre-training and supervised fine-tuning, a financial LLM~(CFLLM) to adeptly manage financial texts, and a deployment framework~(CFAPP) designed to navigate real-world financial applications.
The CFData comprising both a pre-training dataset and a supervised fine-tuning dataset, where the pre-training dataset collates Chinese financial data and analytics, alongside a smaller subset of general-purpose text with 584M documents and 141B tokens in total, and the supervised fine-tuning dataset is tailored for six distinct financial tasks, embodying various facets of financial analysis and decision-making with 1.5M instruction pairs and 1.5B tokens in total.
The CFLLM, which is based on InternLM-7B to balance the model capability and size, is trained on CFData in two stage, continued pre-training and supervised fine-tuning.
The CFAPP is centered on large language models (LLMs) and augmented with additional modules to ensure multifaceted functionality in real-world application.
Our codes are released at \url{https://github.com/TongjiFinLab/CFGPT}\footnote{Correspondence to dcheng@tongji.edu.cn}.

\end{abstract}

\section{Introduction}

In recent years, pre-trained language models (PLMs) have garnered significant attention in both industrial and research arenas, emerging as a key in natural language processing and interactive artificial intelligence~\cite{devlin2018bert, radford2018improving, brown2020language, raffel2020exploring, ouyang2022training}.
The public unveiling of GPT-3.5 and GPT-4 in 2022 marked a significant advancement, as these large language models (LLMs) showcased unparalleled performance in a range of tasks, from reading comprehension and open-ended question answering to code generation~\cite{openai_chatgpt, openai_2023}.
Notably, these LLMs~\cite{openai_chatgpt, du2022glm, openai_2023} exhibit profound competencies in natural language understanding (NLU) and can execute a variety of tasks by following natural language instructions without the need for training data.
Despite these successes, the intricacies of financial texts pose challenges that demand domain-specific LLMs for effective comprehension of sophisticated financial language and concepts.
To this end, several financial large language models (FinLLMs) have been developed, including FinGPT~\cite{yang2023fingpt, zhang2023instruct}, PIXIU~\cite{xie2023pixiu}, and BloombeergGPT~\cite{wu2023bloomberggpt}.

There are also some early endeavors toward the Chinese financial large language model.~\cite{zhang2023self, zhang2023xuanyuan, lu2023bbt, Cornucopia-LLaMA-Fin-Chinese}.
For instance, BBT-FinT5~\cite{lu2023bbt} fine-tuned T5~\cite{raffel2020exploring} using the masked language model task (MLM) and the knowledge-enhanced triple mask task (KETM).
XuanYuan2.0~\cite{zhang2023self, zhang2023xuanyuan}, on the other hand, effectively fine-tuned BLOOM-176B~\cite{scao2022bloom} with both general and domain-specific datasets, showcasing prowess in general and financial question-answering tasks.
Cornucopia~\cite{Cornucopia-LLaMA-Fin-Chinese} further innovated by supervised fine-tuning LLaMA~\cite{touvron2023llama, touvron2023llama2}, enhancing its reasoning abilities within the Chinese financial domain.
Despite the substantial achievements of LLMs in the financial domain, their application to the Chinese financial context not abundantly studied.

In this work, we introduce a Chinese Financial Generative Pre-trained Transformer framework, named CFGPT, including a dataset for pre-training and supervised fine-tuning, a financial LLM to adeptly manage financial texts, and a framework designed to navigate real-world financial applications.
We begin by constructing the CFData, comprising both a pre-training dataset and an supervised fine-tuning dataset.
The pre-training dataset collates an extensive assortment of Chinese financial data and analytics, alongside a smaller subset of general-purpose text.
In aggregate, it contains 584M documents and 141B tokens, encompassing announcements, research reports, social media content, financial news articles, and entries from Wikipedia.
In contrast, the supervised fine-tuning dataset is tailored for six distinct financial tasks, embodying various facets of financial analysis and decision-making.
It consists of 1.5M instruction pairs and 1.5B tokens, covering areas such as sentiment analysis, event detection, report summarization, topic decomposition, question answering, and stock movement prediction.
We chose InternLM-7B as our base model in initial phrase. Our framework also support most of LLM as base model.
Leveraging CFData, our CFLLM undergoes a two-stage training regimen: continued pre-training and supervised fine-tuning.
This approach aims to amplify its zero-shot and few-shot performance in Chinese financial tasks.

We conducted an exhaustive analysis of the critical needs in the Chinese financial domain, leading to the creation of our CFAPP framework.
Designed for real-world applications, this framework is centered on large language models (LLMs) and augmented with additional modules to ensure multifaceted functionality.
A standout feature of CFAPP is its adaptability to diverse input formats—ranging from text and audio to PDF files, enabling users to engage with the system in their preferred manner and enhancing its versatility and user-centricity.
To guarantee the veracity and precision of responses, our framework incorporates several interaction modules: vector databases, a chain-of-thought system, and domain-specific model.
Beyond its input flexibility, CFAPP also offers varied output formats, such as raw text, templated text, and mind maps.
Collectively, these integrative features position our framework as a pioneering solution in the Chinese financial sector, adeptly catering to its unique demands.

\section{Related Work}
In this section, we will discuss related works from three aspects, the financial dataset, the financial language models, the financial evaluation benchmarks. 

\subsection{Financial Dataset}

Financial datasets typically encompass two primary categories: text datasets and structured datasets.
In this section, our attention is centered on the text datasets, specifically, those comprising textual content from the financial sector, devoid of any accompanying tables or figures.
Historically, the primary focus of such datasets has been on news and social media text, offering valuable insights for informed decision-making.
For instance, \citet{zhang2010trading} capitalized on the sentiment discernible in blogs and news articles to predict stock prices.
Similarly, both \citet{ding2014using} and \citet{liang2020f} harnessed news articles, transforming them into structured event relations as a means to predict stock movements.
Further explorations by \citet{liu2021finrl} and \citet{cheng2022financial} delved into extracting latent features from raw textual data, using this auxiliary information to bolster the training of trading systems.

With the popularity of LLMs, there has been a notable shift in research focus, from being model-centric to data-centric, which magnified the importance of textual data within the financial domain.
For instance, BloombergGPT~\cite{wu2023bloomberggpt} integrates both general and financial textual data to train FinLLMs from the ground up.
Similarly, models such as PIXIU~\cite{xie2023pixiu} and FinGPT~\cite{yang2023fingpt, zhang2023instruct} curate instruction datasets.
These datasets reframe tasks like sentiment analysis, named entity detection, and stock price prediction within a question-answering paradigm, facilitating supervised fine-tuning for financial applications.
Focusing on the Chinese financial domain, BBT-Fin~\cite{lu2023bbt} introduced the BBT-FinCorpus, which aggregates roughly 3B tokens from sources like corporate reports, research reports, social media, and financial news.
Meanwhile, XuanYuan 2.0~\cite{zhang2023xuanyuan, zhang2023self} offers a hybrid-tuning dataset combining 380 tokens from both general and financial domains for pre-training and supervised fine-tuning, although this dataset remains proprietary.
Cornucopia~\cite{Cornucopia-LLaMA-Fin-Chinese} emphasizes supervised fine-tuning, constructing a dataset with 26M instruction pairs covering areas such as insurance, investment, stocks, funds, loans, and credit cards.
Recognizing the crucial role of data quality in crafting efficient FinLLMs, we introduce the CFData corpus.
It encompasses 584M documents with 141B tokens for pre-training and an additional 1.5M instruction pairs with 1.5B tokens explicitly designed for supervised fine-tuning.

\begin{table*}[]
\centering
\resizebox{\textwidth}{!}{%
\setlength\tabcolsep{4pt}
\begin{tabular}{lccccccc}
\hline
Dataset                       & \# Docs $\times$10$^3$ & \# Chars $\times$10$^6$ & \# Tokens $\times$10$^6$ & Chars/Doc & Tokens/Doc & \% Token & Storage (GB) \\ \hline
Pretraining                   & 583,978 & 206,665 & 140,609 & 354     & 241     & 100.00 & 573.2 \\ \hline
CFData-CP                & 39.1    & 13,357  & 8,788   & 341,423 & 225,330 & 6.24   & 37.0  \\
CFData-CA                & 6,193   & 31,120  & 17,272  & 5025    & 2789    & 12.28  & 86.3  \\
CFData-RR                & 392     & 5,027   & 3,529   & 12,826  & 9,003   & 2.51   & 13.9  \\
CFData-FN                & 82,438  & 37,262  & 26,297  & 452     & 319     & 18.70  & 103.3 \\
CFData-SM                & 494,661 & 119,708 & 84,587  & 242     & 171     & 60.15  & 332.0 \\
CFData-Wiki              & 255     & 191     & 137     & 750     & 537     & 0.09   & 0.5   \\ \hline\hline
supervised fine-tuning            & 1,572   & 2,042   & 1,512   & 1,299   & 962     & 100.00 & 5.66  \\ \hline
CFData-SA                & 120     & 118     & 86      & 982     & 711     & 5.69   & 0.33  \\
CFData-ED                & 490     & 461     & 343     & 941     & 701     & 22.69  & 1.28  \\
CFData-TD                & 369     & 266     & 187     & 721     & 507     & 12.37  & 0.74  \\
CFData-RS                & 369     & 1,014   & 765     & 2,751   & 2,076   & 50.60  & 2.81  \\
CFData-QA                & 12      & 8       & 6       & 648     & 470     & 0.39   & 0.02  \\
CFData-SP                & 212     & 175     & 125     & 822     & 588     & 8.27   & 0.48  \\ \hline
\end{tabular}
}
\caption{Detailed statistics about the pretraining dataset and the supervised fine-tuning dataset.
\# Docs, \# Chars, and \# Tokens indicate the number of documents, characters and tokens in each sub-dataset.
Chars/Doc, Tokens/Doc indicate the number of characters and tokens per document in each sub-dataset.
\% Token indicates the percentage of the overall tokens of each sub-dataset.
Storage~(GB) indicates the storage occupation of each sub-dataset in terms of gigabyte~(GB).}
\label{table:data_statics}
\end{table*}

\subsection{Financial Language Models}

Before the integration of LLMs into the financial sector, the focus of pre-trained language models largely revolved around continued pre-training using financial texts. The adaptability of BERT~\cite{devlin2018bert} to the financial domain was demonstrated when \citet{araci2019finbert} and \citet{yang2020finbert} pre-trained it on English finance news and communications.
The results significantly surpassed competitive baselines in financial sentiment analysis tasks.
As for Chinese, Mengzi-fin~\cite{zhang2021mengzi} and BBT-FinT5~\cite{lu2023bbt} were trained with analogous tasks and achieved improvements in multiple financial tasks.

The advent of GPT~\cite{openai_chatgpt, openai_2023} spurred a heightened interest in FinLLMs.
BloombergGPT~\cite{wu2023bloomberggpt} stands out as a pioneering FinLLM, boasting 50 billion parameters and trained entirely from scratch.
Its performance metrics, particularly under zero-shot and few-shot scenarios, further underscored the need for domain-specific training.
Successor models like FinGPT~\cite{yang2023fingpt, zhang2023instruct} and PIXIU~\cite{xie2023pixiu} delved deeper into supervised fine-tuning, and their outcomes surpassed BloombergGPT and other generic LLMs~\cite{black2022gpt, du2022glm} in zero-shot or few-shot configurations, thereby accentuating the efficacy of supervised fine-tuning in enhancing in-context learning.
Turning our attention to the Chinese financial domain, models such as XuanYuan2.0~\cite{zhang2023self, zhang2023xuanyuan} showcased prowess in both generic and financial question-answering after fine-tuning on domain-specific datasets.
Cornucopia~\cite{Cornucopia-LLaMA-Fin-Chinese} advanced supervised fine-tuning techniques with LLaMA~\cite{touvron2023llama, touvron2023llama2}, elevating  reasoning capabilities of LLaMA in the Chinese financial domain.

\subsection{Financial Evaluation Benchmarks}

The inception of financial evaluation benchmarks, FLUE, was heralded by \citet{shah2022flue}.
This benchmark comprised five tasks: sentiment analysis~\cite{malo2014good}, news headline classification~\cite{sinha2021impact}, named entity recognition~\cite{alvarado2015domain}, structure boundary detection, and question answering~\cite{maia201818}.
Recognizing a gap for more comprehensive financial applications within FLUE, \citet{xie2023pixiu} augmented the benchmark by adding a stock movement prediction task, resulting in the FLARE benchmark.
As for the Chinese financial domain, \citet{lu2023bbt} pioneered the BBT-CFLEB benchmark, encompassing six tasks: news classification, summarization, relation extraction, question answering, negative news determination, and sentiment analysis.
Recently, \citet{zhang2023fineval} presented FinEval, a benchmark specifically designed for the financial domain knowledge in the LLMs, including 4,661 multiple-choice questions covering finance, economy, accounting, and certificate.

\section{Datasets}

In this section, we present the CFData datasets for pre-training and supervised fine-tuning.

\subsection{Overview}

In this section, we present the CFData for pre-training and supervised fine-tuning.
The CFData can be described from two perspectives:
1) Pre-training Dataset: The dataset encompasses a large amount of Chinese financial data and analytics and a small amount of general-purpose text for pre-training, such as announcements, research reports, social media content, financial news articles, and Wikipedia.
2) supervised fine-tuning Dataset: The dataset includes six financial tasks to reflect different aspects of financial analysis and decision-making, which include sentiment analysis, event detection, report summarization, topic decomposition, question answering, and stock movement prediction.
CFData provides much text information in the financial domain, allowing a FinLLM to learn from different of sources.
In Table~\ref{table:data_statics}, we provide the detailed statistics about our CFData.

\begin{figure*}[t]
\centering
\includegraphics[width=0.96\textwidth]{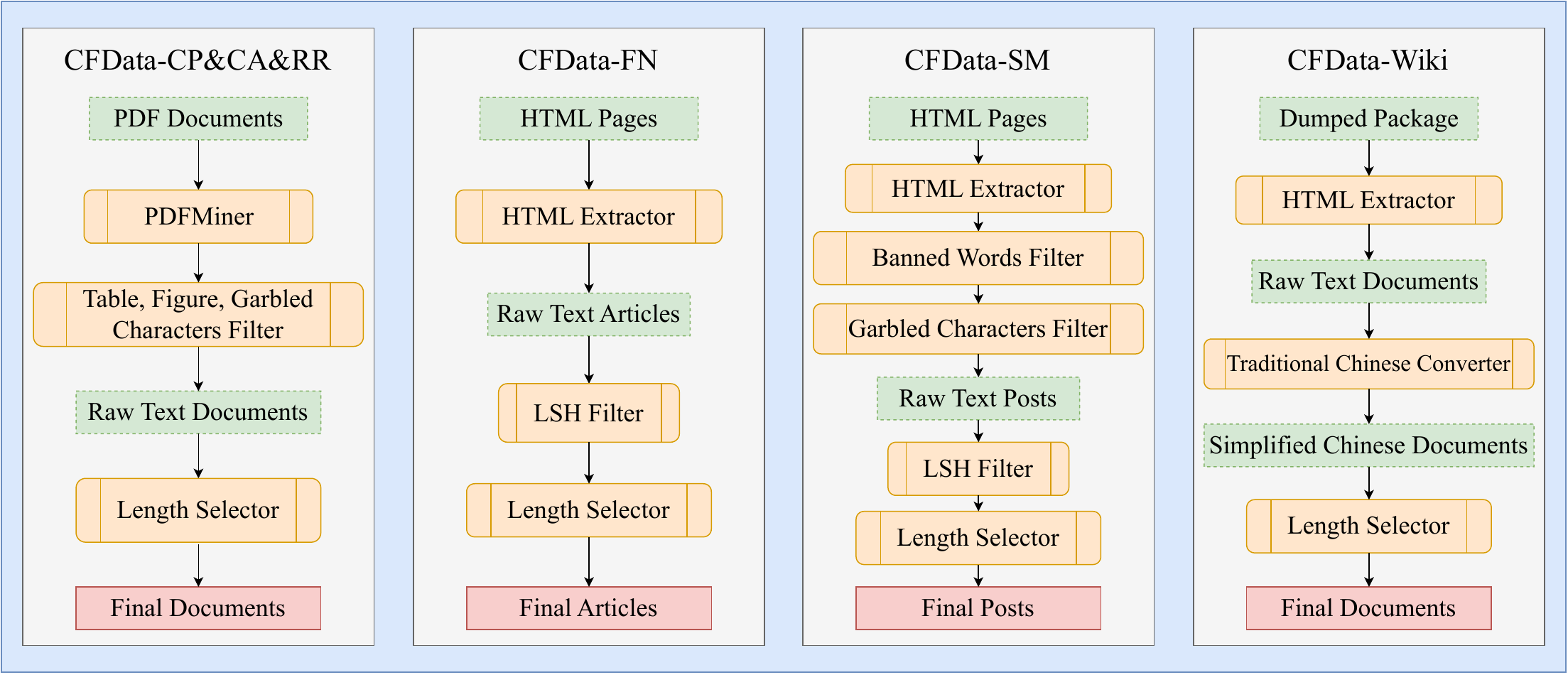}
\caption{The preprocess steps of each sub-dataset in CFData pre-training dataset.}
\label{fig:preprocess}
\end{figure*}

The pre-training dataset consists of 591 million documents and 193 billion tokens, including six sub-datasets,
1) CFData-CP (6.24\%): 39 thousand corporate prospectus with 13 billion tokens;
2) CFData-CA (12.28\%): 6 million corporate announcements with 17 billion tokens;
3) CFData-RR (2.51\% ): 392 thousand research reports with 3 billion tokens;
4) CFData-FN (18.70\%): 82 million financial news with 26 billion tokens;
5) CFData-SM (60.15\%): 495 million social medias and 84 billion tokens;
6) CFData-Wiki (0.09\%): 255 thousand Wikipedia content with 137 million tokens.
More details about the data source and the pre-processing of each sub-dataset are in Section~\ref{sec:data_pretrain} and Figure~\ref{fig:preprocess}.

The supervised fine-tuning dataset consist 1.6 million instructions pairs and 1.5 billion tokens, including six financial tasks,
1) CFData-SA (5.69\% ): 120 thousand instances with 86 million tokens for sentiment analysis;
2) CFData-RS (50.60\%): 369 thousand instances and 765 million tokens for report summary;
3) CFData-ED (22.69\% ): 490 thousand instances with 343 million tokens for event detection;
4) CFData-TD (12.37\%): 369 thousand instances and 187 million tokens for topic decomposition;
5) CFData-QA (0.39\%): 12 thousand instances and 6 million tokens for question-answering;
6) CFData-SP (8.27\%): 212 thousand instances and 125 million tokens for stock moving prediction.
More details about the data source and the task construction of each task are in Section~\ref{sec:data_instruction}.

\subsection{Data for Pre-training}\label{sec:data_pretrain}

In this section, we have gathered additional Chinese financial documents to facilitate pre-training.
This dataset comprises of financial documents, including corporate prospectuses, corporate announcements, research reports, social media content, financial news articles, and the Wikipedia content.
We used a proxy-based distributed crawler to crawl public web pages to get these documents through the API from CFData\footnote{\url{https://github.com/TongjiFinLab/CFGPT-dataset}}.
Moreover, we follow the similar pre-processing steps in \cite{zeng2021pangu} and \cite{lu2023bbt} to clean the dataset.
The inclusion of these documents aims enabling the LLMs to better grasp the intricacies of financial concepts and language within the Chinese context.
Compared with BloombergGPT~\cite{wu2023bloomberggpt}, CFData also consists more long (tokens pre document) financial documents  for pre-training, especially in corporate prospectus, corporate announcements, and research report.

\subsubsection{CFData-CP}
Corporate prospectuses are critical legal documents prepared by companies to provide information to potential investors.
They typically include details about the financial status, business model, strategies, risks, management team, legal proceedings, regulatory compliance, and the utilization of funds by the respective company.
Considered the most important and official document in the financial market, corporate prospectuses play a significant role in investment decisions.
To construct this sub-dataset, we crawl the corporate prospectuses from Shanghai Stock Exchange website\footnote{\url{http://www.sse.com.cn/disclosure/overview/}} and Shenzhen Stock Exchange website\footnote{\url{http://www.szse.cn/disclosure}} spanning the years from 2012 to 2022, where we got 43,371 documents in PDF format initially.
To process the corporate prospectuses for pre-training, we first convert the PDF documents into raw text format through the PDFMiner package\footnote{\url{https://pypi.org/project/pdfminer/}}.
Then, we remove all figures, tables, and garbled characters present in the documents through regular expression.
Additionally, we filter out documents with a length smaller than 10,000 characters.
After these preprocessing steps, the CFData-CP sub-dataset consists of 39.1 thousand corporate prospectuses with a total of 8.79 million tokens.

\subsubsection{CFData-CA}
Corporate announcements are official financial statements released by listed companies to provide information to the general public.
These announcements can be categorized into two types: regular announcements, such as annual and quarterly reports, and irregular announcements that address or disclose unexpected events.
To construct this sub-dataset, we crawl the corporate prospectuses from listed companies in Shanghai Stock Exchange website and Shenzhen Stock Exchange website spanning the years from 2012 to 2022, where we got 9,389,193 documents in PDF format initially.
We following the similar operation as CFData-CP to process the corporate announcements, including, converting the PDF documents into raw text format through the PDFMiner, removing all figures, tables, and garbled characters present in the documents through regular expression, and filtering out documents with a length smaller than 1,000 characters.
Finally, the CFData-CA sub-dataset consists of 6.19 million corporate announcements with a total of 17.3 billion tokens.

\subsubsection{CFData-RR}
Research reports are specialized documents that concentrate on macroeconomic issues, sectors, industries, and stocks.
They are typically issued by investment institutions and aim to analyze the current status and future development trends of the aforementioned topics.
%These reports often reflect the unique perspectives and insights of the analysts, making them valuable for enhancing the reasoning ability of existing large language models.
These reports not only contains the unique perspectives and insights of the analysts but also provide the analyzing step toward their perspectives and insights, making them important to reflect the logic of analysts in financial domain.
To construct this sub-dataset, we crawl the research report from Eastmoney website\footnote{\url{https://data.eastmoney.com/report/}} spanning the years from 2016 to 2022, where we got 683,294 documents in PDF format initially.
We following the similar operation as CFData-CP and CFData-CA to process the research report, including, converting the PDF documents into raw text format through the PDFMiner, removing all figures, tables, and garbled characters present in the documents through regular expression, and filtering out documents with a length smaller than 2,000 characters.
Finally, the CFData-RR sub-dataset consists of 392 thousand research report with a total of 3.53 billion tokens.

\subsubsection{CFData-FN}
Financial news are the news articles related to financial market with financial jargon and acronyms, ranging from stock market updates, company earnings reports, mergers and acquisitions, economic data releases, central bank decisions, government policies, exchange rates, interest rates, and so on.
To construct the sub-dataset with high-quality financial news in terms of factual and unbiased information, we crawl the financial news from reputable Chinese financial websites, including Sina Finance\footnote{\url{https://finance.sina.com.cn/}}, Tencent Finance\footnote{\url{https://new.qq.com/ch/finance/}}, Phoenix Finance\footnote{\url{https://finance.ifeng.com/}}, 36Kr\footnote{\url{https://36kr.com/}}, and Cailianshe\footnote{\url{https://www.cls.cn}}, spanning the years from 2018 to 2022.
We crawl a total of 121,273,632 Chinese financial news in HTML initially, where there are about 61, 66, 63, 68, 71 thousand news pre day in 2018, 2019, 2020, 2021 and 2022, respectively.
To process the financial news for pre-training, we first extract the news articles into raw text through regular expression.
Then we implement the locality-sensitive hashing~\cite{datar2004locality} (LSH) algorithm to filter out the redundant or duplicated news across multiple sources, resulting in a diverse and non-repetitive dataset.
Moreover, we remove the financial news whose length are smaller than 100 characters.
Finally, the CFData-FN sub-dataset consists of 82.4 million news articles, comprising a total of 26.3 billion tokens.

\begin{table*}[]
\centering
\begin{tabular}{ll}
\hline
Task & Instruction Pair                                            \\ \hline
\multirow{3}{*}{CFData-SA}  & Please analyze the sentiment of the following financial paragraph.                       \\
     & The answer should be choose from {``Positive'', ``Negative'', ``Neutral''}.     \\
     & The paragraph is ``[paragraph]''.                                               \\ \hline
\multirow{3}{*}{CFData-ED}                     & Please detect the ``[event category]'' from the following financial paragraph.           \\
     & If the ``[event category]'' exists, find all the event, otherwise, return None. \\
     & The paragraph is ``[paragraph]''.                                               \\ \hline
CFData-RS                       & Please summarize the following financial report. The report is ``[report]''.             \\ \hline
\multirow{2}{*}{CFData-TD} & Please decompose the following financial topic from multiple small aspects.              \\
     & The topic is ``[topic]''.                                                       \\ \hline
\multirow{3}{*}{CFData-QA}  & Please answer the questions based on given financial paragraph and conversation history. \\
                                               & The financial paragraph is ``[paragraph]''. The conversation history is ``[history]''.   \\
     & The question is ``[question]''.                                                 \\ \hline
\multirow{4}{*}{CFData-SP}       & Please analyze the text information and price information of  ``[stock name]'',          \\
     & and determine how will the price change.                                        \\
     & The answer should be choose from {``Positive'', ``Negative'', ``Neutral''}.     \\
     & The text information is ``[text]''. The price information is ``[price]''.       \\ \hline
\end{tabular}
\caption{The instruction pair for each sub-dataset.
The corresponding content are filled into ``{[}{]}'' during supervised fine-tuning.}
\label{table:instruction_example}
\end{table*}

\subsubsection{CFData-SM}
Social media contents are usually posted by individuals to represent their viewpoints, which is quite important for a FinLLM to comprehend the elementary investors in financial market.
There are various platforms for individual investors to discuss the financial market or release the posts, with Xueqiu\footnote{\url{https://xueqiu.com/}}, Guba\footnote{\url{https://guba.eastmoney.com/}}, and Weibo\footnote{\url{http://weibo.com}} being the most active ones.
To construct this sub-dataset, we crawl the posts from these three platforms spanning the years from 2018 to 2022, where only the posts mentioned sectors, industries, stocks, futures, and options will be crawled.
We crawl a total of 20,838,204,735 posts in HTML initially, where there are about 11 millions news pre day.
However, it is important to note that the posts from social media platforms may not always represent reputable sources and could potentially contain biased perspectives or garbled characters.
To address these concerns, we implement certain measures in our data filter process.
\begin{itemize}
    \item We extract posts into raw text through regular expression;
    \item We utilize banned word dictionary from~\cite{zeng2021pangu} to filter out the posts that exhibit toxicity, identity attacks, insults, threats, profanity, and sexually explicit;
    \item We remove posts with garbled characters that comprised more than 30\% of the total characters and eliminate all garbled characters present in the remaining posts;
    \item We implement the locality-sensitive hashing~\cite{datar2004locality} (LSH) algorithm to filter out the duplicated posts.
    \item We filter out posts with a length of less than 50 characters to ensure a minimum level of meaningful content.
\end{itemize}
By applying these filtering criteria, we construct the CFData-SM sub-dataset with 495 million social media posts, comprising a total of 84.6 billion tokens.

\subsubsection{CFData-Wiki}
Wikipedia is a online encyclopedia, containing multiple subjects in multiple areas, which is important to introduce general topics over the work.
To address maintain the generalization ability of FinLLMs during pre-training, we incorporate data from the Chinese Wikipedia page.
To accomplish this, we obtained a dump of the Chinese Wikipedia until July 20, 2023 from the Wikipedia\footnote{\url{https://dumps.wikimedia.org/zhwiki/latest/}}.
To ensure the cleanliness of the dataset for pre-training, we performed several preprocessing steps.
Firstly, we extracted the Chinese documents from the dumped file through regular expression.
Additionally, we converted traditional Chinese characters into simplified Chinese characters to maintain consistency within CFData.
Furthermore, we remove all garbled characters present in each document.
By applying these preprocessing steps, we obtained a final dataset consisting of 255 thousand documents with a total of 137 million tokens.

\subsection{Data for Supervised Fine-tuning} \label{sec:data_instruction}
Derived from real-world finance applications, our supervised fine-tuning dataset is constructed using open-sourced data from six distinct tasks: financial sentiment analysis, financial event detection, financial report summary, financial topic decomposition, financial question answering, and stock movement prediction.
The specific prompts for each task can be found in Table~\ref{table:instruction_example}.

\subsubsection{CFData-SA}
Financial sentiment analysis is a crucial task in the financial domain.
While there are existing datasets available, most of them are in English.
To enhance the FinLLM specifically for financial sentiment analysis in Chinese market, we construct the supervised fine-tuning dataset in two aspects,
1) label the social media posts from the CFData-SM by the GPT-4 API;
2) correlate the content and invest rating in the research report from the CFData-RR.

For the first approach, we first random select 60 thousand social media posts with the length more than 100 characters, where we select 12 thousand social media posts each year from 2018 to 2022.
Then we explore the GPT-4 API to label the sentiment of each post with ``Positive'', ``Negative'', and ``Neutral'', where the prompt can be found in Table~\ref{table:instruction_example}.
For the second approach, we first random select 60 thousand research reports with the length between 2000 to 3000 characters, where we select 8.57 thousand research reports each year from 2016 to 2022.
Then we extract the content and the invest rating from each research report, where the content is regarded as the text condition, and the invest rating is mapped to ``Positive'', ``Negative'', and ``Neutral'' and regarded as the target.
This approach allows us to capture the sentiment within the specific content of the report, considering the unique characteristics of the Chinese financial market.
By combining these two approaches, we construct the CFData-ED task comprising 120 thousand items with a total of 118 million tokens.

\subsubsection{CFData-ED}
Financial event detection aims to categorize the financial documents with a hierarchy is helpful for people to make informed investment decisions in the financial market~\cite{liang2020f}.
Given a financial document, \emph{e.g.}, news, research reports and announcements of listed companies, the financial event is detected based on a tree-structured event scheme, which contains 98 event categories spreading across the event category nodes of the constructed event category tree of depth 7.
Specifically, we explore the CN-Fin~\cite{liang2020f} dataset to construct the financial event detection task, where we randomly select 490 thousand instances out of all 500 thousand instance to construct CFData-ED dataset, and leave 10 thousand instances for future benchmark construction.
By regarding the event detection as multi-label classification with the prompt in Table~\ref{table:instruction_example}, we construct the CFData-ED task comprising 490 thousand items with a total of 461 million tokens.

\subsubsection{CFData-TD}
In real-world financial applications, decomposing a general question into multiple simple or concrete questions is helpful to proceed the search and summary towards the original question, which is also important for retrieval augmented generation (RAG) chatbot to answer the question thoroughly with more evidence.
For example, one might inquire about the development of electric vehicle industry in China.
However, answering such questions requires breaking them down into multiple aspects to gather qualitative and quantitative information, which can then be used to address the original questions effectively.
The logical decomposition of the topic into different aspects is crucial for constructing a dataset that supports this type of analysis.

To accomplish this without extensive manual labor, we propose leveraging research reports, as their titles typically contain the topic of interest, while their outlines reflect the analytical approach used to study that topic.
To enhance a FinLLM in the context of financial topic decomposition, we propose constructing the dataset by using the title of each research report as the source topic and the outline of each research report as the decomposed target.
This dataset will enable the model to learn how to decompose financial topics into their constituent aspects.
By filtering the research report without outlines or titles from CFData-RR, we get 379 thousand research reports in total, where we randomly select 369 thousand research to construct CFData-TD dataset and leave 10 thousand instances for future use.
Then we split the titles and outlines of each research report to construct this dataset.
Finally, we construct the CFData-TD task comprising 369 thousand items with a total of 187 million tokens.

\subsubsection{CFData-RS}
Report summary holds significant importance in financial applications.
Given that financial reports have specific focus areas distinct from general documents, enhancing a FinLLM for financial report summary is of utmost importance.
To accomplish this without extensive manual labor, we propose to leveraging the research report itself to provide supervision.
In detail, we construct a this dataset by considering the content of the research report as the source and utilizing the conclusion and abstract of the research report as the target.
This approach allows us to capture the key information and main points from the research reports, facilitating the generation of concise and informative summaries.
Following the similar split in CFData-TD, we select 369 thousand research to construct CFData-RS dataset and leave 10 thousand instances for future use.
By implementing this methodology, we construct the CFData-RS task comprising 369 thousand items with a total of 765 million tokens.

\begin{figure}[t]
\centering
\includegraphics[width=0.6\columnwidth]{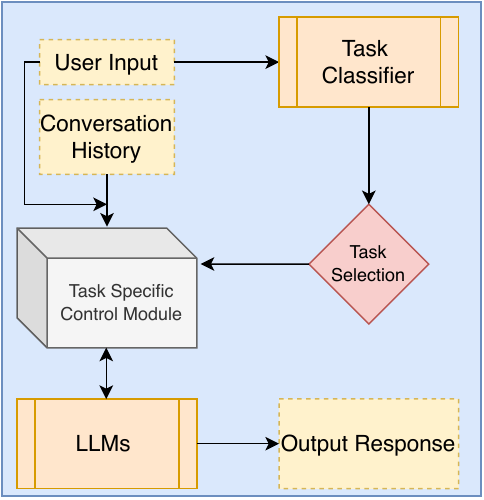}
\caption{The working pipeline of CFAPP framework}
\label{figure:pipeline}
\end{figure}

\begin{figure}[t]
\centering
\includegraphics[width=0.9\columnwidth]{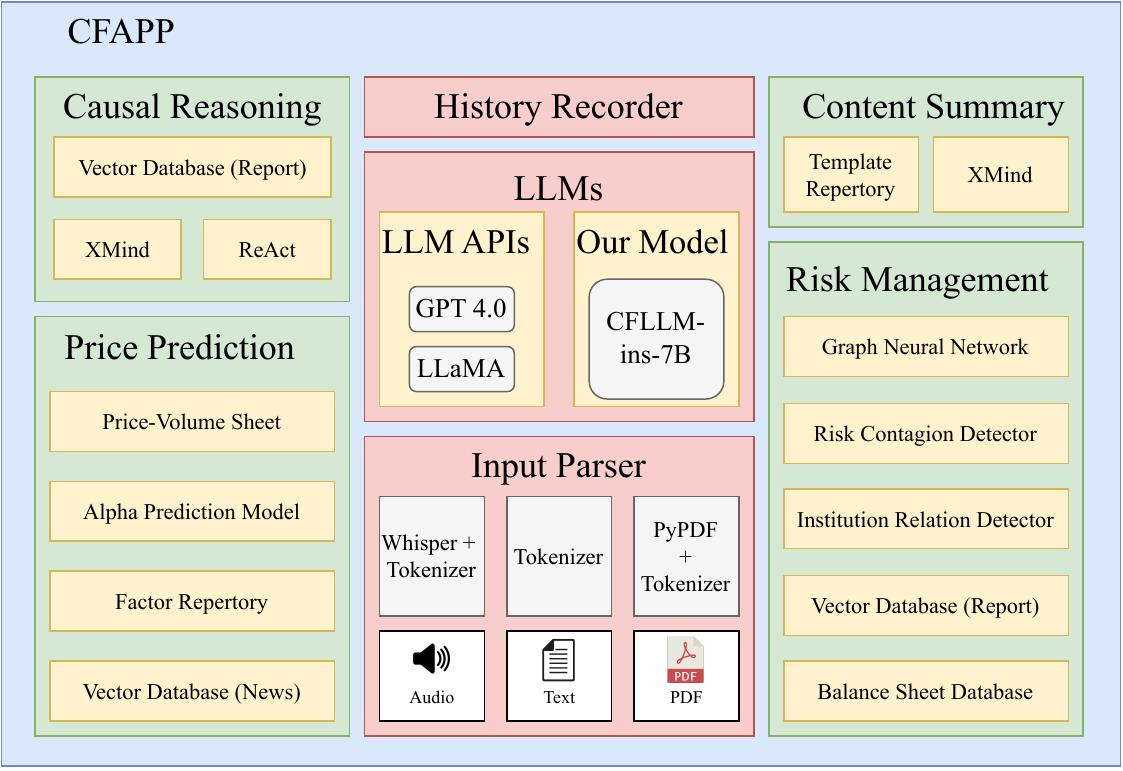}
\caption{The CFAPP framework}
\label{figure:CFAPP}
\end{figure}

\subsubsection{CFData-QA}
Question answering is an important task that involves automatically providing answers to financial questions based on the conditional information.
However, the majority of existing financial document question-answering datasets are in English \cite{chen2021finqa, chen2022convfinqa}.
To address this limitation and enhance the performance of our CFGPT model in financial question answering, we propose translating two existing English financial question-answering datasets: FinQA \cite{chen2021finqa} and ConvFinQA \cite{chen2022convfinqa}, into Chinese.
FinQA comprises question-answering pairs annotated by experts, along with corresponding earnings reports from S\&P 500 companies.
ConvFinQA builds upon FinQA by introducing multi-turn conversations for question-answering tasks using earnings reports.
By merging these two translated datasets, we successfully construct the CFData-QA task comprising 12 thousand items with a total of 6 million tokens.

\subsubsection{CFData-SP}
Stock moving prediction is a fundamental financial task with significant potential value in real-world applications, particularly in quantitative investment.
In line with the formulation in BigData22~\cite{soun2022accurate}, we construct the CFData-SP dataset by combining the most popular social media posts, financial news, and historical stock prices to forecast future stock price movements.
In detail, we first select the mostly viewed related financial news and related posts, historical close price in past five day, and close price changing rate in next day for each stock in CSI Smallcap 500 Index and each trading day from 2021 to 2022.
Then we frame the task as a classification problem, with the goal of predicting whether the stock close price in next day will ascend, descend, or hold.

Specifically, we employ the following classification criteria:
1) If the close price changing rate exceeds 0.50\%, the sample is classified as ``Ascend'';
2) If the close price changing rate falls below -0.50\%, the sample is classified as ``Descend'';
3) Otherwise, samples are categorized as ``hold''.
Therefore, we can capture the different directions and magnitudes of stock price movements in our dataset.
Finally, we construct the CFData-SP task comprising 212 thousand items with a total of 175 tokens.

\section{Model and Training}

Our CFLLM model is built based on the base model, InternLM-chat-7b.
To enhance the understanding and reasoning abilities of base model within the financial domain, and align the base model to real-world financial application, we formulate the fine-tuning process into two stage: continued pre-training and supervised fine-tuning.

\subsection{Continued Pre-training}\label{sec:continue_pretrain}
In the first stage, we explore the InternLM-chat-7b as the base model, which is fine-tuned with a standard left-to-right causal language modeling objective on the our pre-training dataset in Sec.\ref{sec:data_pretrain}.
To maximize GPU utilization, we follow \cite{brown2020language} and cut all our training sequences to be the same length, in our case 1,024 tokens.
To achieve this, we first concatenate all our tokenized training documents with an ``$<$EOS$>$'' token as a document separator.
Then we generate the the training instance with sample gap and sample length as 512 and 1,024, respectively.
Note that, the InternLM-chat-7b model explore the relative positional encoding and FlashAttention, our model can also be applied to sequence longer than 1,024 during the inference.
During the continued pre-training, we use the AdamW optimizer~\cite{loshchilov2017decoupled}, with the $\beta_1$, $\beta_2$, and weight decay as 0.9, 0.95, and 1e-5 respectively.
The batch size is set as 512, learning rate is set as 1$\times$10$^{-5}$, and we use the cosine decay learning rate scheduler with linear warmup in the first 1000 steps.
The continued pre-training is execute on 8 pieces of A800 80GB GPUs.
After the continued pre-training, we get our CFLLM-pt-7B model.

\subsection{Supervised Fine-tuning}
In the second stage, we explore the continued pretrained model in Sec~\ref{sec:continue_pretrain} as the base model, which is fine-tuned with QLoRA~\cite{hu2021lora} in supervised fine-tuning data covering 6 financial tasks in Sec.~\ref{sec:data_instruction} and Moss-03-sft dataset~\cite{sun2023moss} to balance the general response ability and domain specific ability of our model.
To further enhance the ability of model dealing with long financial documents, we set the maximum length of input texts as 2048.
To balance the GPU utilization and training instance length, we sorted all training instance to make the training instance within each batch have the similar sequence, and the batch size is between 64 to 512, which is decided by the sequence length of each batch.
During the supervised fine-tuning, we use the AdamW optimizer~\cite{loshchilov2017decoupled}, with the $\beta_1$, $\beta_2$, and weight decay as 0.9, 0.95, and 1e-5 respectively.
The learning rate is set as 2$\times$10$^{-4}$, and we use the cosine decay learning rate scheduler with linear warmup in the first 500 steps.
For the QLoRA setting, we set the LoRA rank as 64, the LoRA Alpha as 16 and the dropout rate of the LoRA linear function was set to 0.05.
The supervised fine-tuning is execute on 8 pieces of A800 80GB GPUs.
After the supervised fine-tuning, we get our CFLLM-ins-7B model.

\begin{figure*}[t]
\centering
\includegraphics[width=0.96\textwidth]{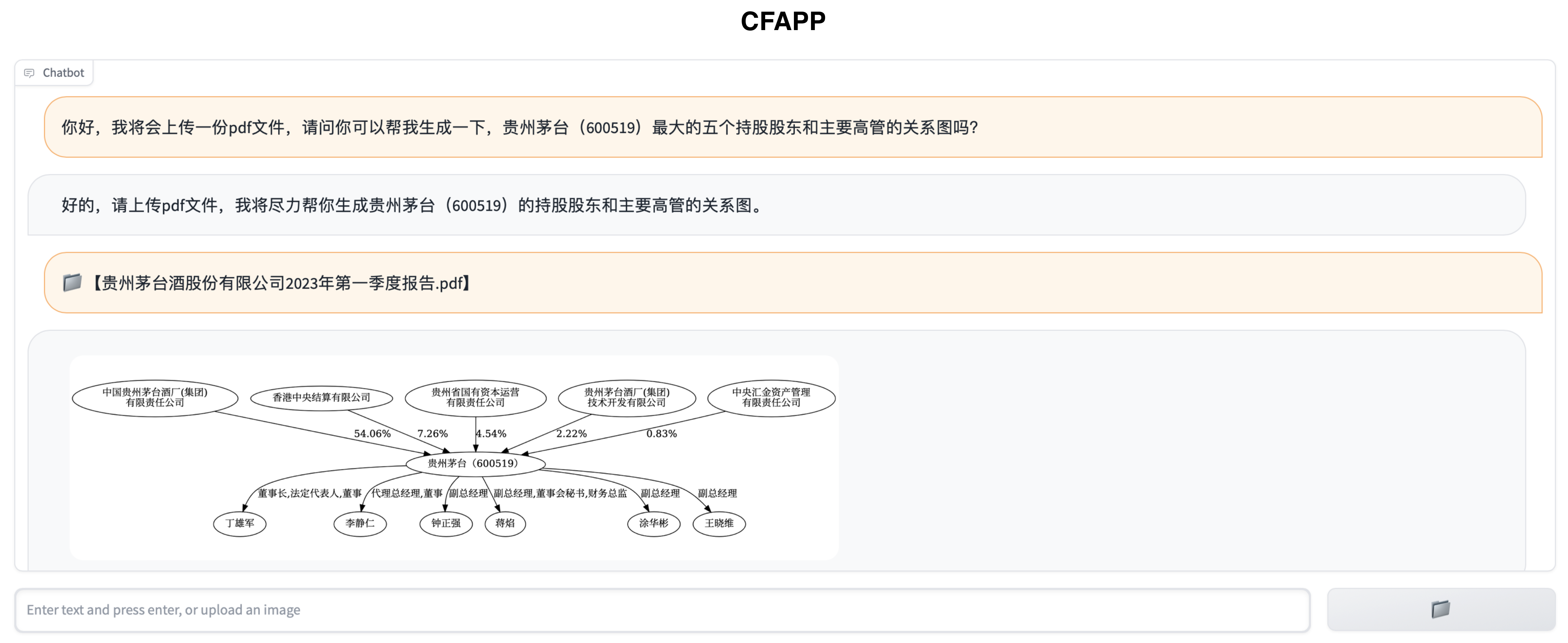}
\caption{The example of our \textit{Content Summary} to summarize the shareholding relationship and executive chart based on the corporate announcements.}
\label{figure:demo_mindmap}
\end{figure*}

\section{Application}
In this section we will introduce our CFAPP framework and provide examples to show the capability our demonstration.

\subsection{Systemic Description}

In Figure~\ref{figure:pipeline}, we introduce the working pipeline of our CFAPP framework.
Specifically, given a user input, we first explore a task classifier to identify the the functional task of the input, then we seed the user input along with the conversation history to the task specific control module, where we designed different working procedure to each of the task.
For each task specific control module, it will interact with the LLMs multiple times to get the output response.
In Figure~\ref{figure:CFAPP}, we introduce all the components of our CFAPP framework, where the \textit{LLMs} provide multiple choice to different large language models, the \textit{Input Parser} provides three different apis for different kind of inputs, the \textit{History Recorder} stores the conversation history in previous iteration, the \textit{Content Summary}, \textit{Causal Reasoning}, \textit{Price Prediction}, and \textit{Risk Management} contains the model, database, and tools for their corresponding requirements.
Note that the \textit{Content Summary} and \textit{Causal Reasoning} functionalities mainly focus on enhance the representation and reasoning ability of the LLMs in financial field, while the \textit{Price Prediction} and \textit{Risk Management} functionalities pay more attention to incorporate the task-specific model with LLMs to enhance the justification and decision ability of our system.
In the following sections, we will introduce the detailed implementation of each functionality.

\subsubsection{Content Summary}

The content summary feature of our system allows users to emphasize specific documents or paragraphs of interest.
Our system supports multiple output formats, including template summaries, mind map summaries, and text summaries.
When interacting with our system, users can indicate their desired output format, and we provide support for generating summaries in these multiple formats.
For content summarization, we utilize the conversation history and the user input to seed the large language model (LLM) and generate item-wise summaries and the response types.
If the response type is ``template'', we retrieve the corresponding template from our template repository and combine it with the item-wise summary to generate the response.
If the response type is ``mind map'', we directly generate a mind map based on the item-wise summary using the graphviz package.
If the response type is ``raw text'', we seed the item-wise summary back into the LLM to generate a coherent and comprehensive paragraph as the response.
By offering these different response types, we provide users with flexibility in selecting the output format that best suits their needs and preferences.
This allows users to obtain summaries in the format that they find most useful and convenient.

\subsubsection{Causal Reasoning}

Our framework includes a causal reasoning component that enables users to answer complex questions with the support of evidence.
It is important to note that our framework supports both conditional question-answering and open-domain question-answering.
For open-domain question-answering, the user inputs the question along with ReAct format prompts (which involve the loop of Thought, Action, and Observation) to the large language model (LLM).
The ReAct prompts guide our framework to retrieve relevant documents from the vector database as supporting evidence.
The ReAct prompts allow for two types of actions: ``[search]'' and ``[response]''.
If the action is ``[search]'', our framework retrieves the top-5 relevant documents from the vector database.
On the other hand, if the action is ``[response]'', the LLM directly generates the answer based on the user input, conversation history, and the retrieved documents.
After generating the response, we present all the ``Thought'' and the answer to the users, demonstrating how our framework decomposes and solves the question.
For conditional question-answering, the user provides both the question and the condition to the LLM, along with revised ReAct prompts.
The revised ReAct prompts include three types of actions: ``[search]'', ``[response]'', and ``[pass]''. The ``[pass]'' action is used when the Thought can be captured within the given condition.
The revised ReAct prompts allow the LLM to determine whether the condition is sufficient to generate the answer.
If the condition is enough to answer the question, our framework directly generates the answers.
However, if the condition is not sufficient, a similar procedure is executed as in the open-domain question-answering scenario.
Through these mechanisms, our framework enables users to address both conditional and open-domain questions, leveraging evidence from the vector database and the reasoning capabilities of the LLM.
This provides users with a robust and comprehensive approach to answering complex questions.

\subsection{Demonstration Example}

\begin{figure*}[t]
\centering
\includegraphics[width=0.96\textwidth]{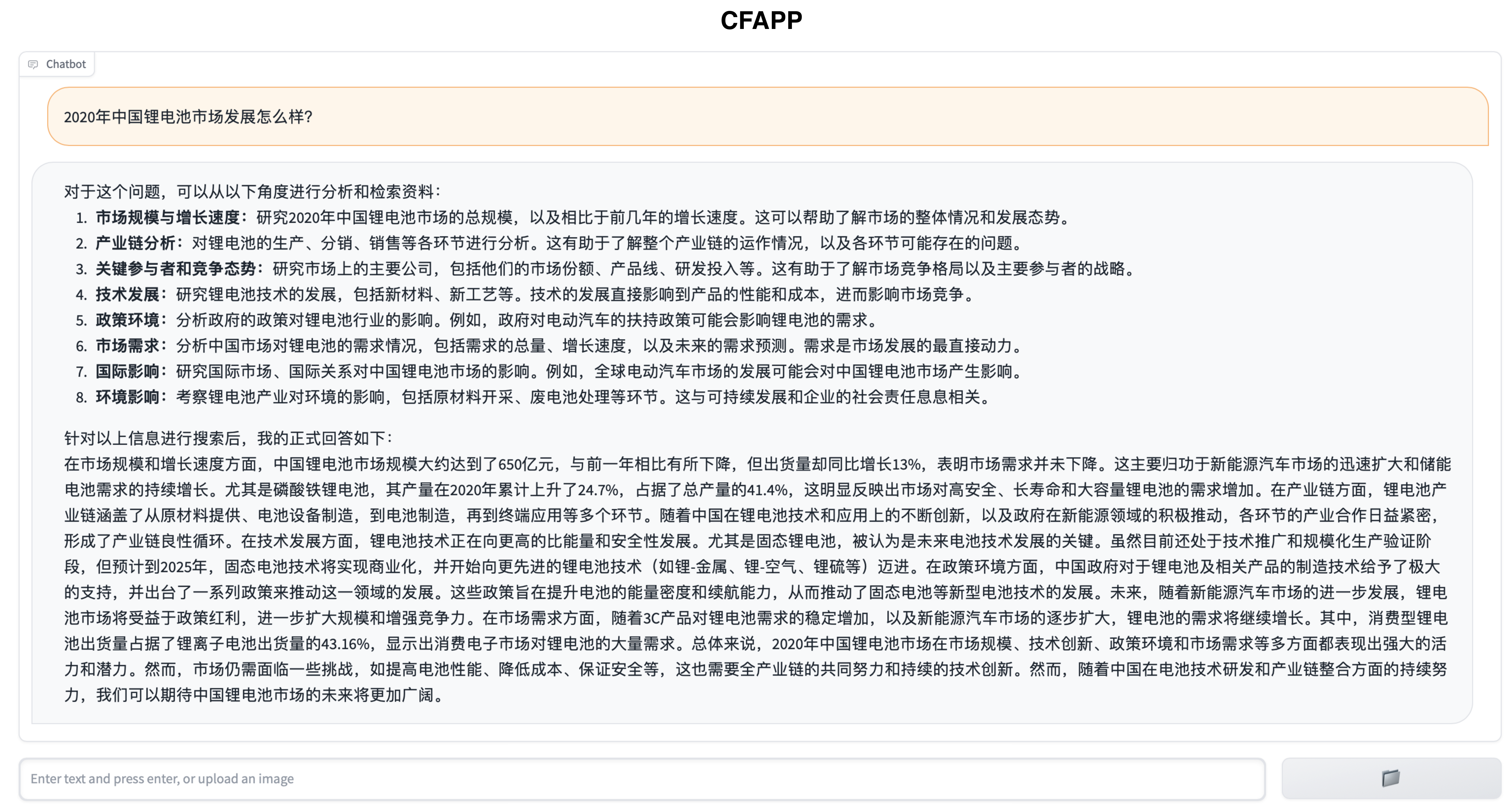}
\caption{The example of our \textit{Causal Reasoning} to answer the open-domain question.}
\label{figure:demo_react}
\end{figure*}

In Figures~\ref{figure:demo_mindmap} and \ref{figure:demo_react}, we show two examples of our framework to reveals the functionality of \textit{Content Summary} and \textit{Causal Reasoning}.

In the first example from Figure~\ref{figure:demo_mindmap}, we provide the framework with a PDF document related to the industrial structural of a listed company ``Kweichow Moutai Company Limited (SH:600519)'', and then the users asks to summary the shareholding relationship and executive chart of Kweichow Moutai Company with mind map, where our model can accomplish this work precisely with the mind map as output.
The detailed conversation is translated as follows:
\begin{itemize}
    \item User: Hello, I will upload a PDF file, can your help to summary the top-5 shareholding relationship and executive chart of Kweichow Moutai (600519)?
    \item CFAPP: Yes, please upload the PDF file, I will try to generate the shareholding relationship and executive chart of Kweichow Moutai (600519).
    \item User: Upload a PDF file, named as ``\textit{Kweichow Moutai Co., Ltd. First Quarter Report for the year 2023}''
    \item CFAPP: A mind map about the top-5 shareholding relationship and top-6 executive chart of the Kweichow Moutai (600519).
\end{itemize}

In the second example from Figures~\ref{figure:demo_react}, we ask a open-domain question, ``'', and our framework can return the question decomposition procedure and the answer together as the response to show how our framework analyze this open-domain question and then interact with the vector dataset to answer question with evidence.
The detailed conversation is translated as follows:
\begin{itemize}
    \item User: How did China's lithium battery market develop in 2020?
    \item CFAPP: The question should be considered in the following aspects: 1. Market Size and Growth Rate: Research the overall scale of China's lithium battery market in 2020 and its growth rate compared to previous years. This can help understand the overall market situation and development trend; 2. Industry Chain Analysis: Analyze the production, distribution, and sales of lithium batteries at each stage. This helps understand the operation of the entire industry chain and identify potential issues at each stage; 3. Key Participants and Competitive Landscape: Research the major companies in the market, including their market share, product lines, and research and development investments. This helps understand the market's competitive landscape and the strategies of key participants; 4. Technological Advancements: Research the development of lithium battery technology, including new materials, processes, and innovations. Technological advancements directly impact product performance and costs, which in turn influence market competitiveness; 5. Policy Environment: Analyze the impact of government policies on the lithium battery industry. For example, government support policies for electric vehicles may influence the demand for lithium batteries; 6. Market Demand: Analyze the demand for lithium batteries in the Chinese market, including the total demand, growth rate, and future demand forecasts. Demand is the most direct driving force for market development; 7. International Impact: Study the influence of the international market and international relations on the Chinese lithium battery market. For instance, the development of the global electric vehicle market may have an impact on China's lithium battery market; 8. Environmental Impact: Investigate the environmental effects of the lithium battery industry, including raw material extraction and waste battery disposal. This is closely related to sustainable development and corporate social responsibility. \\
        Based on the search results towards these question, our answer can be organized as
        In terms of market size and growth rate, the Chinese lithium battery market reached approximately 65 billion yuan in 2020, experiencing a slight decline compared to the previous year. However, the shipment volume increased by 13\% year-on-year, indicating that market demand remained robust. This can be attributed to the rapid expansion of the new energy vehicle market and the continuous growth in demand for energy storage batteries. Notably, the production of lithium iron phosphate batteries increased by 24.7\% in 2020, accounting for 41.4\% of the total production, which reflects the increasing demand for high safety, long-life, and high-capacity lithium batteries in the market.Regarding the industry chain, the lithium battery industry covers multiple segments, including raw material supply, battery equipment manufacturing, battery production, and end applications. With China's continuous innovation in lithium battery technology and applications, as well as the government's active promotion in the new energy field, cooperation across various segments of the industry chain has become increasingly closer, creating a positive cycle within the industry chain.In terms of technological development, lithium battery technology is progressing towards higher specific energy and safety. Particularly, solid-state lithium batteries are considered a key technology for future battery development. Although currently in the stage of technology promotion and scaled production verification, it is expected that by 2025, solid-state battery technology will be commercialized and begin to move towards more advanced lithium battery technologies, such as lithium-metal, lithium-air, and lithium-sulfur batteries.Regarding the policy environment, the Chinese government has provided strong support for lithium battery manufacturing technology and has implemented a series of policies to drive the development of this field. These policies aim to improve battery energy density and range, thereby promoting the development of new battery technologies like solid-state batteries. In the future, with further development in the new energy vehicle market, the lithium battery market will benefit from policy dividends, leading to further expansion in scale and enhanced competitiveness.Regarding market demand, with the steady increase in demand for lithium batteries from 3C products (computers, communications, and consumer electronics) and the gradual expansion of the new energy vehicle market, the demand for lithium batteries will continue to grow. Consumer-grade lithium batteries account for 43.16\% of the total lithium-ion battery shipments, highlighting the significant demand from the consumer electronics market.Overall, the Chinese lithium battery market showed strong vitality and potential in 2020 across various aspects such as market size, technological innovation, policy environment, and market demand. However, the market still faces some challenges, such as improving battery performance, reducing costs, and ensuring safety. These challenges require collaborative efforts and continuous technological innovation across the entire industry chain. Nevertheless, with China's ongoing efforts in battery technology research and industry chain integration, we can expect a promising and broader future for the Chinese lithium battery market.
\end{itemize}

\section{Conclusion}

In this study, we introduced an open-sourced Chinese financial assistant with large language model, named CFGPT, for applicable LLM requirements in the financial domain. CFGPT comprises a deployment framework to cater to real-world applications, a two-stage fine-tuned open-source LLM on Chinese financial datasets, and a benchmark for evaluating model performance in the Chinese financial domain.
Through initial evaluation, we showcased the effectiveness of our CFLLM-ins-7B model across financial tasks, underscoring the potential of domain-specific continued pre-training and supervised fine-tuning of large language models in the financial domain.
Nonetheless, challenges such as enhancing performance on complex tasks and addressing resource constraints still exist.
Our open-source contribution aims to foster further research and innovation in financial large language model literature, promoting the development of more valuable and applicable LLMs in the finance sector.

\bibliography{aaai24}

\end{document}